\documentclass[final,3p,times,twocolumn]{elsarticle}

\usepackage{blindtext, graphicx, amsmath, algorithm, algpseudocode, pifont, algcompatible, comment, layout, amsthm, amssymb}
\usepackage{enumitem}   
\usepackage{eso-pic}
\usepackage{booktabs}
\usepackage{float}
\usepackage{longtable}

\usepackage[utf8]{inputenc}
\usepackage[english]{babel}
\usepackage{hyperref} 
\hypersetup{ colorlinks=true, linkcolor=black, filecolor=black, urlcolor=cyan, }

\usepackage{caption}
\usepackage{lineno}
\usepackage{textcomp}
\usepackage{stfloats}
\usepackage{url}
\usepackage{verbatim}
\usepackage{comment}
\usepackage{soul}
\usepackage{xcolor}
\usepackage[acronym, nohypertypes={acronym}, nonumberlist]{glossaries}
\makeglossaries
\usepackage{tikz}
\usetikzlibrary{decorations.pathmorphing}

\journal{ICT Express}
\newacronym{lam}{LAM}{Large AI Model}

\newacronym{llm}{LLM}{Large Language Model}

\newacronym{ai}{AI}{Artificial Intelligence}

\newacronym{ntn}{NTN}{Non-Terrestrial Networks}

\newacronym{drl}{DRL}{Deep Reinforcement Learning}

\newacronym{rl}{RL}{Reinforcement Learning}

\newacronym{wf}{WF}{Water-Filling}

\newacronym{iot}{IoT}{Internet of Things}

\newacronym{leo}{LEO}{Low Earth Orbit}

\newacronym{mdp}{MDP}{Markov decision Process}

\newacronym{ecef}{ECEF}{Earth-centred Cartesian}

\newacronym{sinr}{SINR}{Signal-to-Interference-Plus-Noise Ratio}

\newacronym{kpi}{KPI}{Key Performance Indicators}

\newacronym{pc}{PC}{Proportional Capacity}

\newacronym{mmf}{MMF}{Max–Min Fairness}

\begin{document}

\begin{frontmatter}

\title{Large Artificial Intelligence Model–Guided Deep Reinforcement Learning for Resource Allocation in Non-Terrestrial Networks}
\author{Abdikarim Mohamed Ibrahim 1\corref{cor1}}
\ead{abdikarimi@sunway.edu.my}

\author{Rosdiadee Nordin 2}
\ead{rosdiadeen@sunway.edu.my}

\address{Faculty of Engineering, Sunway University, Petaling Jaya, Malaysia\\Future Cities Research Institute, Sunway University, Petaling Jaya, Malaysia}

\cortext[cor1]{Abdikarim Mohamed Ibrahim}

\begin{abstract}
Large AI Model (LAM) have been proposed to applications of Non-Terrestrial Networks (NTN), that offer better performance with its great generalization and reduced task specific trainings. In this paper, we propose a Deep Reinforcement Learning (DRL) agent that is guided by a Large Language Model (LLM). The LLM operates as a high level coordinator that generates textual guidance that shape the reward of the DRL agent during training. The results show that the LAM-DRL outperforms the traditional DRL by 40\% in nominal weather scenarios and 64\% in extreme weather scenarios compared to heuristics  in terms of throughput, fairness, and outage probability.
\end{abstract}


\end{frontmatter}


\section{Introduction}
\label{sec:introduction}

\acrfull{ntn} inherently have dynamic complexities such as rapid changes in network topologies, heterogeneous user distributions, and changing propagation conditions~\cite{ref3NN}, which present challenges to traditional optimization approaches. As constellations change and expand to accommodate thousands of satellites (e.g., Starlink has already launched 7,213 satellites by October 30, 2024 toward a planned total of 42,000)~\cite{ref1N}, intelligent and adaptive resource management solutions become essential to ensure efficient spectrum utilization and enable consistent service quality across diverse regions~\cite{ref2}.

Traditional resource allocation approaches face three main challenges in such environments. First, classical optimization methods such as \acrfull{wf} and static rule-based models are designed for quasi-static channels and struggle to react to fast topology changes when \acrfull{leo} satellites move at about $(7.5)$~km/s~\cite{ref3}. Second, heuristic approaches face stability issues as the number of beams and user terminals grows, for example when frequent satellite handovers must be handled jointly with power and bandwidth decisions. Third, although \acrfull{drl} has been shown to provide stability in dynamic wireless networks~\cite{ref4N}, \acrshort{drl} agents act as black boxes in the sense that humans cannot easily understand how the neural network maps inputs to actions or justify the selected decisions (i.e., actions)~\cite{ref5}. In addition to this black-box nature, \acrshort{drl} tends to be sample inefficient and requires extensive retraining when network conditions change~\cite{ref6}. Recent work has therefore explored hybrid intelligent schemes, in which generative models or \acrshort{llm}s guide \acrshort{drl} in wireless systems~\cite{ref6,ref7N}. These studies highlight that model-based guidance can improve robustness and interpretability, but they largely focus on terrestrial, low-altitude scenarios, or small AI models such as TinyML \cite{ref8N}, and do not target \acrshort{ntn} specific propagation, heterogeneous user regions, or explicit \acrshort{kpi} driven control.

The goal is to overcome the shortcomings of the traditional \acrshort{drl} (e.g., sample inefficiency and frequent retraining~\cite{ref6}). To achieve this, a prompt is sent to an  \acrshort{llm} that contains the current \acrshort{ntn} state and service providers objectives, and the \acrshort{llm} responds with a strategy (e.g., efficiency-focused, fairness-focused, or high-latitude-priority) for a \acrshort{drl} agent to follow during learning. Then, this response (i.e., strategy) is embedded to the agent learning via the reward. The resulting \acrshort{lam}-\acrshort{drl} agent learns policies that are sample efficient and can be interpreted or explained in terms of strategy usage and feature importance.


The rest of the paper is organized as follows. Section~\ref{sec:related} presents related work on \acrshort{drl}-based \acrshort{ntn} optimization. Section~\ref{sec:system_model} presents the \acrshort{ntn} system model for satellite resource allocation. Section~\ref{sec:approach} presents the proposed \acrshort{llm}-\acrshort{drl} framework. Section~\ref{sec:results} presents performance evaluations. Section~\ref{sec:conclusion} concludes the paper.


\section{Related Works}
\label{sec:related}


\acrshort{drl} has been applied to resource allocation in \acrshort{ntn} and has been shown to outperform the traditional rule based heuristics. For example, Birabwa \emph{et al.}~\cite{ref9N} proposed multi-agent \acrshort{drl} for joint user association and resource allocation in integrated terrestrial and \acrshort{ntn}, and showed that \acrshort{drl} adapts better to rapid topology changes and heterogeneous links compared to rule based heuristics. Similarly, Hu \emph{et al.}~\cite{ref10N} proposed a \acrshort{drl} scheme for multi dimensional resource allocation in \acrshort{leo} satellite uplinks, that jointly controls power and bandwidth in order to improve spectral efficiency under dynamic traffic and mobility. These works demonstrate the potential of \acrshort{drl} for \acrshort{ntn} resource management, but they also highlight training complexity and scalability challenges when network size and dynamics increase.

\acrshort{llm}s have been proposed as promising tools for wireless network optimization and management. Hang \textit{et al.}~\cite{ref7N} reviewed how \acrshort{llm}s can support next-generation networking technologies and argued that they can help address the black-box nature and lack of interpretability inherent in traditional \acrshort{drl}. Sun \textit{et al.}~\cite{ref6} presented a framework that improves traditional \acrshort{drl} from both the data and policy perspectives, using diffusion models and Transformers to refine decision making. Their findings highlighted sample inefficiency as a major bottleneck, since the \acrshort{drl} agent requires extensive interaction with the environment.

However, these studies focus mainly on terrestrial or low altitude scenarios and do not account for the high mobility propagation and complex topology constraints that are unique to \acrshort{leo} \acrshort{ntn}. This paper differs in two main ways. First, we formulate an \acrshort{mdp} for \acrshort{ntn} resource allocation that incorporates ITU-R based propagation and heterogeneous user regions, and we benchmark against both \acrshort{drl} baseline and traditional resource allocation schemes. Second, instead of letting the \acrshort{llm} output resource allocations directly, we use it to generate human interpretable strategies that are embedded into the \acrshort{drl} policy through attention modulation and strategy dependent reward shaping. In this way, the \acrshort{drl} agent remains the main decision maker that learns from interactions, while the \acrshort{llm} provides semantic guidance that improves interpretability and helps address \acrshort{drl} sample inefficiency and retraining limitations.



\section{System Model}
\label{sec:system_model}

\begin{figure}[t]
    \centering
    \includegraphics[width=0.7\linewidth]{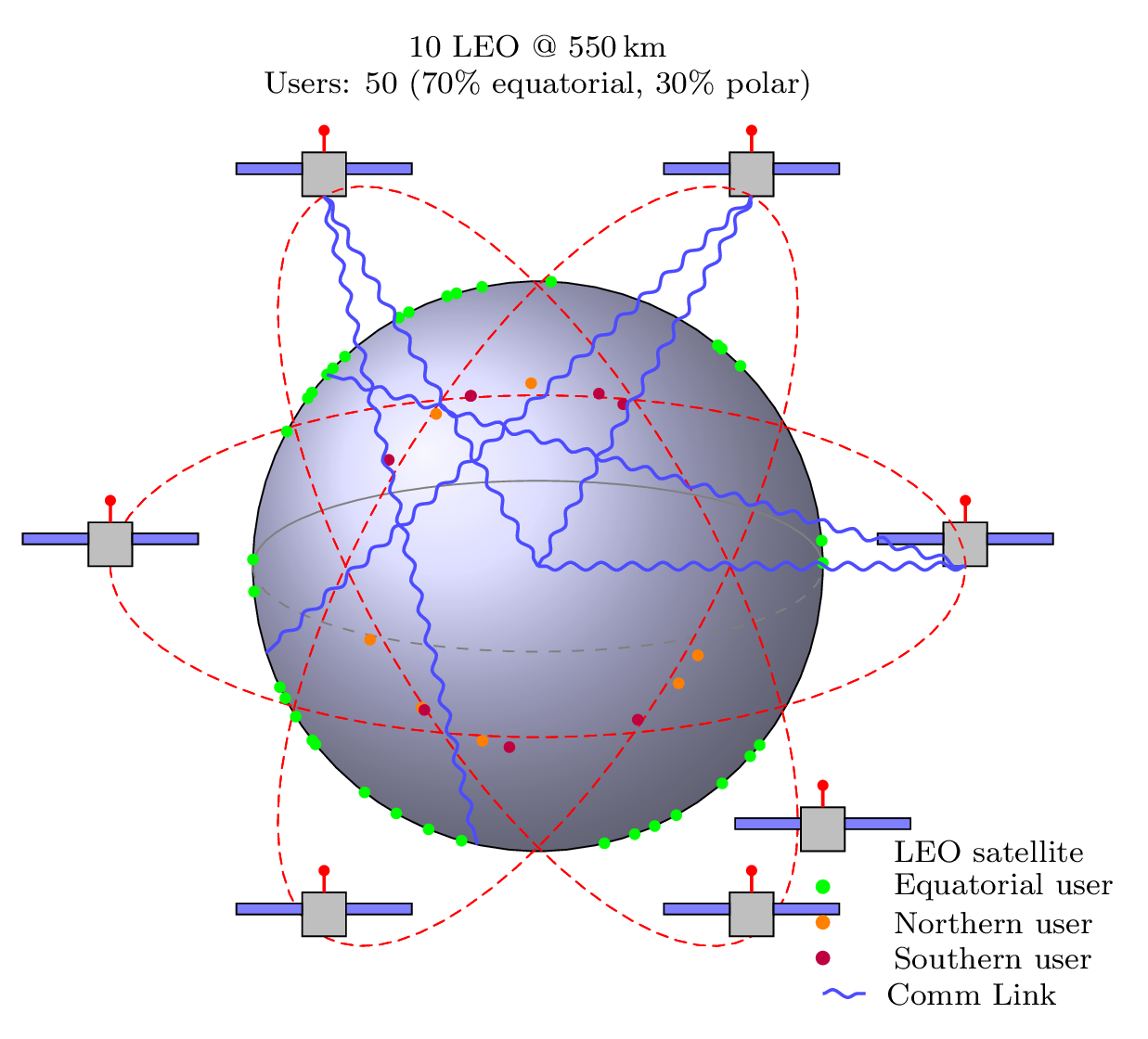}
    \caption{LEO satellite constellation and user distribution across latitude zones.}
    \label{fig:system_model}
\end{figure}

We consider a downlink \acrshort{ntn} with $N_s = 10$ \acrshort{leo} satellites at altitude $h = 550~\mathrm{km}$ and inclination $53^\circ$, that serve $N_u = 50$ ground users. Users are distributed in three latitude zones as follows: $70\%$ in equatorial regions (latitudes between $-30^\circ$ and $30^\circ$), $15\%$ in northern high latitudes ($30^\circ$–$70^\circ$), and $15\%$ in southern high latitudes ($-70^\circ$–$-30^\circ$). Each user is associated with the satellite that provides the largest  average received power.

Satellite orbits follow a Keplerian model with equal spaces of right ascension of
the ascending node. Let $\mathbf{x}_s(t)$ and $\mathbf{x}_u$ represent the positions of
satellite $s$ and user $u$ at time $t$ in an Earth centred frame, and let
$d_{s,u}(t) = \|\mathbf{x}_s(t)-\mathbf{x}_u\|$ be the slant distance.
User positions are fixed within an episode, while $d_{s,u}(t)$ evolves as
satellites move, with the geometry being updated every $30$~s. We operate in Ku band
with carrier frequency $f = 12~\mathrm{GHz}$.

The path loss $L_{s,u}(t)$ \,[dB] combines free–space loss at $f$,
gaseous and rain attenuation from ITU-R P.618-14 and P.676-13, and a fixed
implementation margin~\cite{ref16}. We assume $L_{s,u}(t)$ as the total channel loss between satellite $s$ and user $u$. Each user is served by a spot beam with EIRP cap $P_{\max}$ and bandwidth cap $B_{\max}$. The \acrshort{drl} agent selects normalized
fractions $\alpha_u(t), \beta_u(t) \in [0,1]$, which are mapped to
$P_{t,u}(t) = \alpha_u(t) P_{\max}$ and $B_u(t) = \beta_u(t) B_{\max}$.
We set $P_{\max} = 40~\mathrm{dBm}$ and $B_{\max} = 20~\mathrm{MHz}$ per user.
Let $I_u(t)$ denote the aggregate co–channel interference at user $u$,
$N_0$ which is the noise spectral density, and $G_t$ and $G_r$ the effective
transmit and receive antenna gains are included into the EIRP budget. The
instantaneous \acrfull{sinr} for user $u$ is
\begin{equation}
\gamma_u(t) =
\frac{P_{t,u}(t)\,10^{(G_t + G_r - L_{s(u),u}(t))/10}}
     {I_u(t) + N_0 B_u(t)}.
\label{eq:sinr}
\end{equation}

The instantaneous downlink rate of user $u$ is
\begin{equation}
R_u(t) = B_u(t) \log_2 \big( 1 + \gamma_u(t) \big) \; [\mathrm{bits/s}],
\label{eq:rate}
\end{equation}
and the system sum rate is $R_{\mathrm{sum}}(t) = \sum_{u=1}^{N_u} R_u(t)$. Fairness is measured by Jain's index
\begin{equation}
J(t) =
\frac{\left( \sum_{u=1}^{N_u} R_u(t) \right)^2}
     {N_u \sum_{u=1}^{N_u} R_u^2(t)},
\label{eq:jain}
\end{equation}
which lies in $[0,1]$, with $J(t)=1$ for equal rates. Outage probability is the fraction of users whose \acrshort{sinr} is below a
decoding threshold $\gamma_{\mathrm{th}}$,
\begin{equation}
P_{\mathrm{out}}(t) =
\frac{1}{N_u}
\sum_{u=1}^{N_u}
\mathbf{1} \big( \gamma_u(t) < \gamma_{\mathrm{th}} \big),
\label{eq:outage}
\end{equation}
where $\mathbf{1}(\cdot)$ is the indicator function. We set $\gamma_{\mathrm{th}} = 10^{-3/10} \approx -3~\mathrm{dB}$. This
corresponds to the minimum \acrshort{sinr}, in which a low–order modulation and coding scheme (e.g., QPSK with code rate $1/2$) can maintain a block error rate on the order of $10^{-1}$–$10^{-2}$ on Ku–band satellite links. Therefore, links operating below this level will require a very strong coding or repeated retransmissions and are therefore are treated as being in outage.


\section{Hybrid Intelligence Framework with LAM Integration}
\label{sec:approach}

The \acrshort{ntn} resource allocation problem is modeled as an \acrshort{mdp} guided by an \acrshort{llm} that generates high level strategies to improve the agent's leaning. In the \acrshort{mdp}, the \acrshort{drl} agent observes state $s_t$ that consists of geometry, path loss, and user \acrshort{kpi}s. The agent then, selections action $a_t$ that represents power and bandwidth fractions, and receives a scalar reward $r_t$. The \acrshort{llm} provides a a strategy label $\sigma_t \in \mathcal{S}$. This label is then embedded into the \acrshort{drl} agent reward. We write the MDP as $(\mathcal{S}, \mathcal{A}, P, r, \gamma)$ with action space $\mathcal{A} = [0,1]^{2N_u}$ for the normalized allocations $\{\alpha_u(t), \beta_u(t)\}$, transition dynamics $P(s_{t+1} \mid s_t, a_t)$ affected by orbital motion, traffic, and discount factor $\gamma \in (0,1)$. The total reward at time $t$ is

\begin{equation}
r_t = r(s_t, a_t, \sigma_t)
=
r_t^{\mathrm{base}}(s_t, a_t)
+
\phi_{\sigma_t}(t),
\label{eq:lamdrl-reward}
\end{equation}
where $r_t^{\mathrm{base}}$ combines the three \acrshort{kpi}s,
\begin{equation}
r_t^{\mathrm{base}} =
\lambda_R \frac{R_{\mathrm{sum}}(t)}{R_{\mathrm{ref}}}
+ \lambda_J J(t)
- \lambda_O P_{\mathrm{out}}(t),
\label{eq:base-reward}
\end{equation}
and $\phi_{\sigma_t}(t)$ is a strategy dependent shaping term.
Let $\sigma_t \in \{A,B,C,D\}$ denote the current strategy, and let
$\mathcal{U}_{\mathrm{eq}}$ and $\mathcal{U}_{\mathrm{hl}}$ be the set of equatorial and high–latitude users. Define
\begin{equation}
R_{\mathrm{eq}}(t) = \sum_{u \in \mathcal{U}_{\mathrm{eq}}} R_u(t),
\qquad
R_{\mathrm{hl}}(t) = \sum_{u \in \mathcal{U}_{\mathrm{hl}}} R_u(t),
\label{eq:region-rates}
\end{equation}
and the rate variance
\begin{equation}
V_R(t) =
\frac{1}{N_u} \sum_{u=1}^{N_u}
\big( R_u(t) - \bar{R}(t) \big)^2,
\quad
\bar{R}(t) = \frac{R_{\mathrm{sum}}(t)}{N_u}.
\label{eq:rate-variance}
\end{equation}
The shaping term then is
\begin{equation}
\phi_{\sigma_t}(t) =
\begin{cases}
\eta_{\mathrm{A}} \dfrac{R_{\mathrm{eq}}(t)}{R_{\mathrm{sum}}(t) + \varepsilon},
 & \text{if } \sigma_t = \text{A (equatorial priority)},\\[1.0ex]
-\,\eta_{\mathrm{B}} \dfrac{V_R(t)}{\bar{R}^2(t) + \varepsilon},
 & \text{if } \sigma_t = \text{B (fairness focused)},\\[1.0ex]
\eta_{\mathrm{C}} \dfrac{R_{\mathrm{hl}}(t)}{R_{\mathrm{sum}}(t) + \varepsilon},
 & \text{if } \sigma_t = \text{C (high-latitude priority)},\\[1.0ex]
0, & \text{if } \sigma_t = \text{D (opportunistic efficiency)},\\
\end{cases}
\label{eq:shaping-term}
\end{equation}
with small coefficients $\eta_{\mathrm{A}},\eta_{\mathrm{B}},\eta_{\mathrm{C}} > 0$
and $\varepsilon$ to avoid division by zero.
Therefore, $r_t^{\mathrm{base}}$ in \eqref{eq:base-reward} ensures a global trade–off between the throughput, fairness, and outage.
$\phi_{\sigma_t}(t)$ in \eqref{eq:shaping-term} redistributes the emphasis across user regions according to the \acrshort{llm}–provided intent.

We use an off–policy actor–critic algorithm with continuous actions (i.e., TD3), in which the actor and critic networks are extended with
strategy conditioned attention and the reward in \eqref{eq:lamdrl-reward}. Algorithm~\ref{alg:lamdrl} summarizes the proposed framework.

\begin{figure}[t]
\centering
\includegraphics[width=0.99\linewidth]{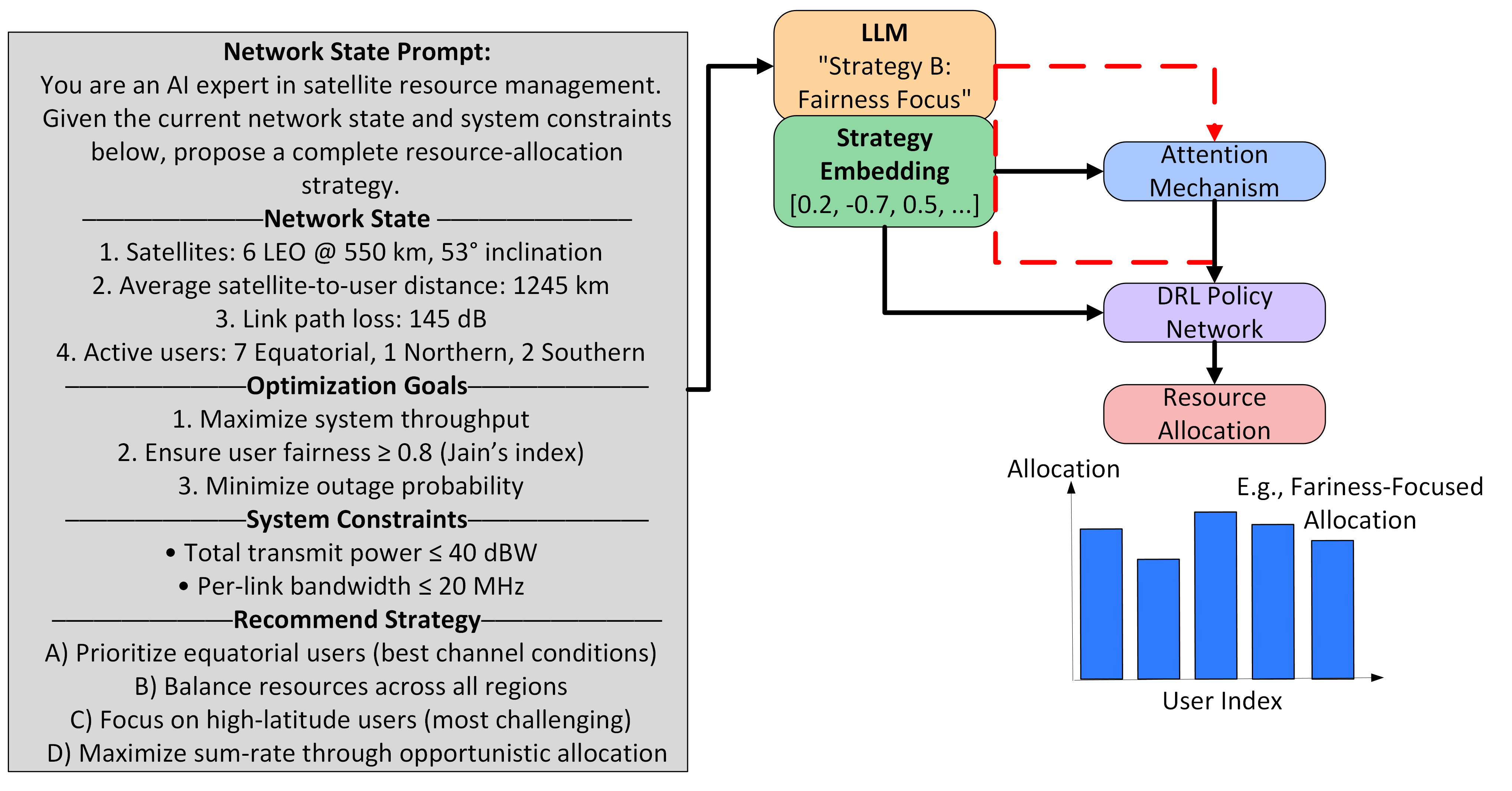}
\caption{Proposed framework. The environment provides a state $s_t$. A prompt summarizes $s_t$ and operator objectives and is sent to the LLM. The  LLM responds with a strategy label $\sigma_t$. The strategy is embedded as a vector $e_{\sigma_t}$ that conditions a single–head attention layer and shapes the reward. 
intent.}
\label{fig:framework}
\end{figure}

\subsection{State Representation and Strategy Generation}

The state $s_t$ collects user–level and satellite–level features. For each user $u$ we include latitude and longitude, propagation features (i.e., slant distance and path loss from Section~\ref{sec:system_model}), a region indicator (i.e., equatorial, northern high latitude, or southern high latitude),
and recent rate $R_u(t-1)$ and \acrshort{sinr} $\gamma_u(t-1)$. These form a feature vector $x_u(t) \in \mathbb{R}^{d_f}$. Putting all user features and global aggregates (i.e., sum rate, Jain index, and outage probability) gives
\[
s_t = [x_1(t), \dots, x_{N_u}(t), g(t)],
\]
where $g(t)$ denotes the global \acrshort{kpi} vector.

At the start of each episode, a prompt is built using aggregates from  $\mathcal{P}$, from the current weather scenario (i.e., nominal or extreme), from user distribution statistics, from the mean and variance of path loss, and from a textual description of the operator intent (i.e., fairness, efficiency, or challenging user coverage). The \acrshort{llm} responds with a label $\sigma \in \mathcal{S} = \{A,B,C,D\}$, which is fixed for that
episode. The \acrshort{llm} therefore observes only the summarized indicators and objectives, and does not output power or bandwidth values.

\subsection{Strategy Embedding and Attention Mechanism}

The strategy $\sigma$ is mapped to a learnable embedding using
$E \in \mathbb{R}^{|\mathcal{S}| \times d_{\mathrm{str}}}$, where
$d_{\mathrm{str}}$ is the embedding dimension. This can be written as
$e_{\sigma} = E(\sigma) \in \mathbb{R}^{d_{\mathrm{str}}}$. For each time step $t$, the user features are $\{ x_u(t) \}_{u=1}^{N_u}$ with
$x_u(t) \in \mathbb{R}^{d_f}$. An additive attention layer computes  context vector $c_t$ conditioned on $e_{\sigma}$:
\begin{equation}
a_u(t) = \mathbf{v}^\top
\tanh \big( W_x x_u(t) + W_e e_{\sigma} \big),
\label{eq:attn-score}
\end{equation}
\begin{equation}
\alpha_u(t) =
\frac{\exp(a_u(t))}
     {\sum_{j=1}^{N_u} \exp(a_j(t))},
\label{eq:attn-weights}
\end{equation}
\begin{equation}
c_t = \sum_{u=1}^{N_u} \alpha_u(t)\, x_u(t),
\label{eq:attn-context}
\end{equation}
where $W_x \in \mathbb{R}^{d_h \times d_f}$,
$W_e \in \mathbb{R}^{d_h \times d_{\mathrm{str}}}$, and
$\mathbf{v} \in \mathbb{R}^{d_h}$ are learned parameters.
The actor network receives $z_t = [c_t,\, g(t),\, e_{\sigma}]$, and outputs the action $a_t = \pi_\theta(z_t) \in [0,1]^{2N_u}$, in which Gaussian exploration noise is added to the training. The critic uses the same attention mechanism to build $c_t$ and approximates $Q_\psi(s_t, a_t, e_{\sigma})$. Actor and critic parameters $(\theta,\psi)$ are updated by TD3, and target
networks and mini batches are sampled from the replay buffer.

\begin{algorithm}[t]
\caption{LAM–DRL Algorithm for LEO NTN Resource Allocation}
\label{alg:lamdrl}
\begin{algorithmic}[1]
\State Initialize actor $\pi_\theta$, critic $Q_\psi$, target networks, replay buffer $\mathcal{D}$
\For{each episode}
  \State Sample user layout and weather scenario
  \State Build prompt $\mathcal{P}$ from aggregated \acrshort{kpi}s and operator intent
  \State Query \acrshort{llm} with $\mathcal{P}$ to obtain strategy label $\sigma$ and embedding $e_{\sigma}$
  \For{$t = 1$ to $T$}
    \State Observe state $s_t = [x_1(t),\dots,x_{N_u}(t),g(t)]$
    \State Compute attention context $c_t$ using \eqref{eq:attn-score}–\eqref{eq:attn-context}
    \State Select action $a_t = \pi_\theta([c_t,g(t),e_{\sigma}]) + \text{noise}$
    \State Execute $a_t$; observe $s_{t+1}$ and instantaneous \acrshort{kpi}s
    \State Compute $r_t^{\mathrm{base}}$ via \eqref{eq:base-reward}, $\phi_{\sigma}(t)$ via \eqref{eq:shaping-term}, set $r_t$
    \State Store $(s_t,a_t,r_t,s_{t+1},e_{\sigma})$ in replay buffer $\mathcal{D}$
    \State Sample mini–batch from $\mathcal{D}$; update critic $Q_\psi$ and actor $\pi_\theta$ with TD3
  \EndFor
\EndFor
\end{algorithmic}
\end{algorithm}

\begin{figure*}[t]
    \centering
    \includegraphics[width=0.8\linewidth]{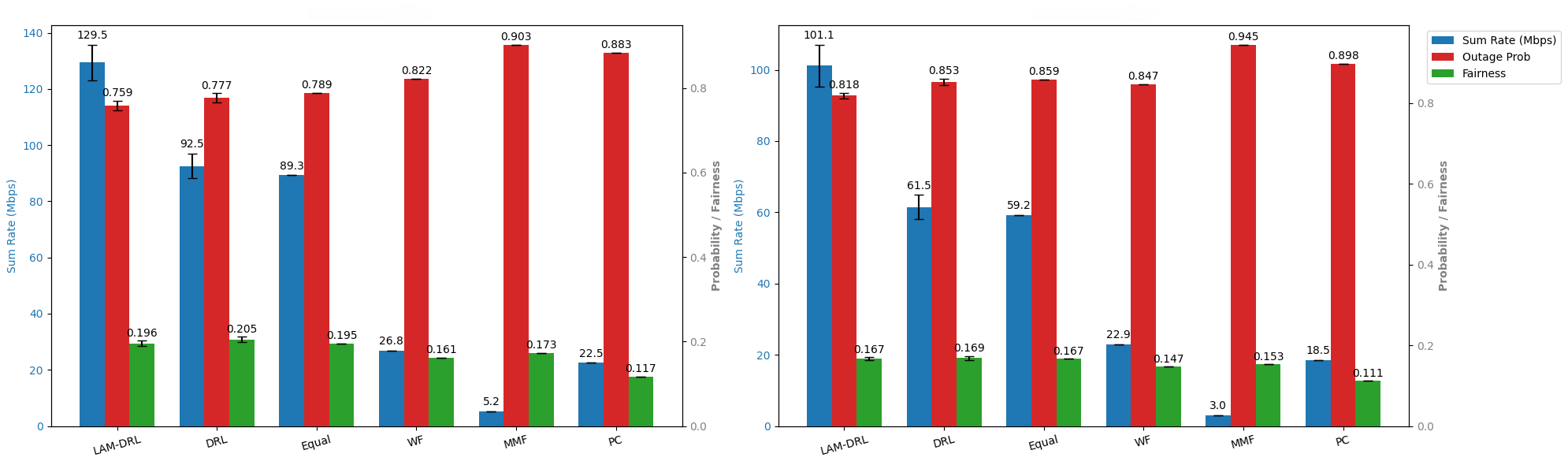}
    \caption{Performance of \acrshort{llm}–\acrshort{drl} and baseline schemes under nominal and extreme weather conditions.}
    \label{fig:lamdrl_results}
\end{figure*}

\section{Performance Evaluation}
\label{sec:results}

In this section, we evaluate the proposed framework. Two weather scenarios are considered namely, nominal and extreme. Nominal scenario refers to a weather with clear sky and light rain, and the extreme scenario refers to a weather with strong rain and high gas absorption. The DRL agent operates under the defined beam caps $P_{\max}$ and $B_{\max}$. At each decision step, we calculate ser rates $R_u(t)$, sum rate $R_{\mathrm{sum}}(t)$, outage probability $P_{\mathrm{out}}(t)$, and Jain’s fairness index $J(t)$ according to \eqref{eq:rate}–\eqref{eq:outage}. A link is considered to be in outage if it is below the decoding threshold $\gamma_{\mathrm{th}} \approx -3~\mathrm{dB}$.

We compare the proposed framework with five main traditional and intelligent schemes as follows: a) the traditional black-box  \acrshort{drl} that has the same actor-critic but without \acrshort{llm} guided strategies; b) the heuristic equal allocation of power and bandwidth for all available users; c)  the traditional  classical water-filling (\acrshort{wf}) resource allocation \cite{ref17}; d) the heuristic max–min fairness (\acrshort{mmf}), which aims to cater for the most disadvantaged users until the allocated budgets are finished \cite{ref18}; and e) the heuristic  proportional capacity (\acrshort{pc}), which allocates resources in proportion to estimated link capacities \cite{ref19}. Learning agents share the same environment and constrains. During evaluation, an episode covers a training window which includes satellite positions, slant ranges, and rain attenuation that are updated every $30$~s. For each weather scenario, we run $100$ independent test episodes and report empirical means with one standard deviation. The \acrshort{llm} is queried once at the beginning of each episode to select the strategy.

Figure~\ref{fig:lamdrl_results} presents the results for both weather scenarios. We observe that, in the nominal case, the \acrshort{llm}–\acrshort{drl} achieves an average sum rate of $129.5$~Mbps, which is around $40\%$ higher than black-box \acrshort{drl} and is far better in performance compared to the heuristic schemes. It also maintains fairness around $J \approx 0.76$ and a lower outage probability. Another observation is that the outage probability is high for all approaches. This is because the combination of the large slant ranges, Ku-band operation, and rain attenuation is strong and adds large-scale losses per-user caps $P_{\max}=40$~dBm and $B_{\max}=20$~MH, which makes edge users to always fall below the decoding threshold $\gamma_{\text{th}}$. Despite this, we can observe that the \acrshort{llm}–\acrshort{drl} consistently achieves lower outage than the baselines by reallocating resources toward users that are falling below the threshold and almost being in an outage. Similarly, in the extreme weather scenario, all approaches degrade due to stronger attenuation, but \acrshort{llm}–\acrshort{drl} is still the most robust scheme that delivers $64\%$ higher sum rate than traditional \acrshort{wf} and maintains higher fairness compared to heuristic baselines.

\begin{figure}[t]
    \centering
    \includegraphics[width=0.9\linewidth]{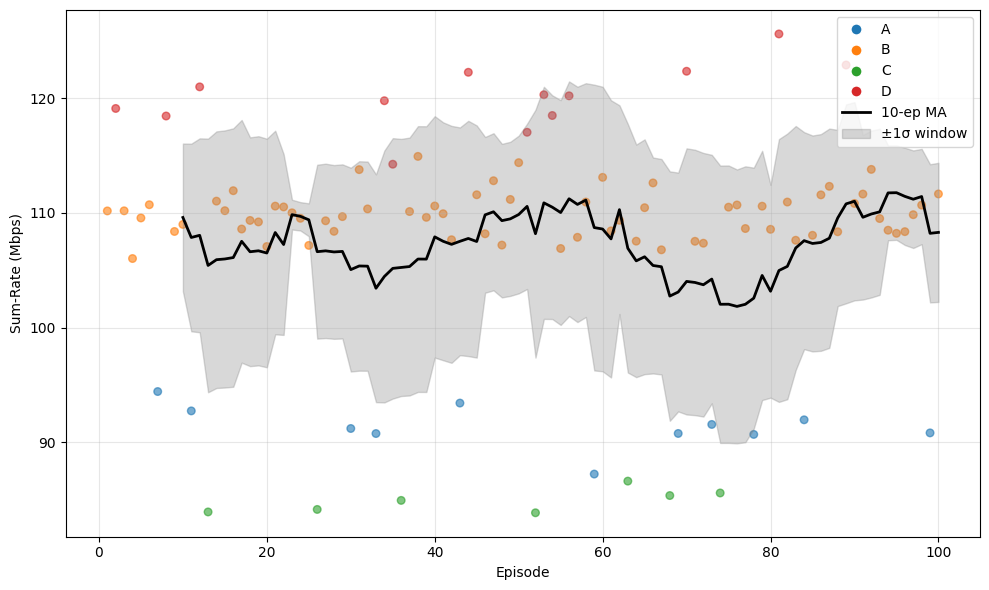}
    \caption{Strategy usage and associated sum-rate performance for \acrshort{llm}–\acrshort{drl} across episodes. Points show per-episode sum rate coloured by the selected \acrshort{llm} strategy; the solid curve is a 10-episode moving average with a shaded $\pm 1\sigma$ window.}
    \label{fig:strategy_dist}
\end{figure}

In order to understand the impact of the textual guidance by the \acrshort{llm} on the \acrshort{drl} agent learning, we log the strategies selected by the \acrshort{llm} at the start of each episode along with its sum rate. Figure Figure~\ref{fig:strategy_dist} shows strategy selection, and we note that in early episodes, all four strategies (i.e., A: efficiency, B: fairness focus, C: challenging-user priority, D: opportunistic allocation) are being selected with the same frequency, which explains the high variance in the performance. However, with more training and learning time, strategy~B becomes dominant and most selected and the moving average of $R_{\text{sum}}$ stabilizes at a higher level with reduced variance. This shows that the agent is able to exploit the \acrshort{llm} guidance (i.e., strategies) and establish that a fairness policy simultaneously improves throughput and outage.

\begin{figure}[t]
    \centering
    \includegraphics[width=0.99\linewidth]{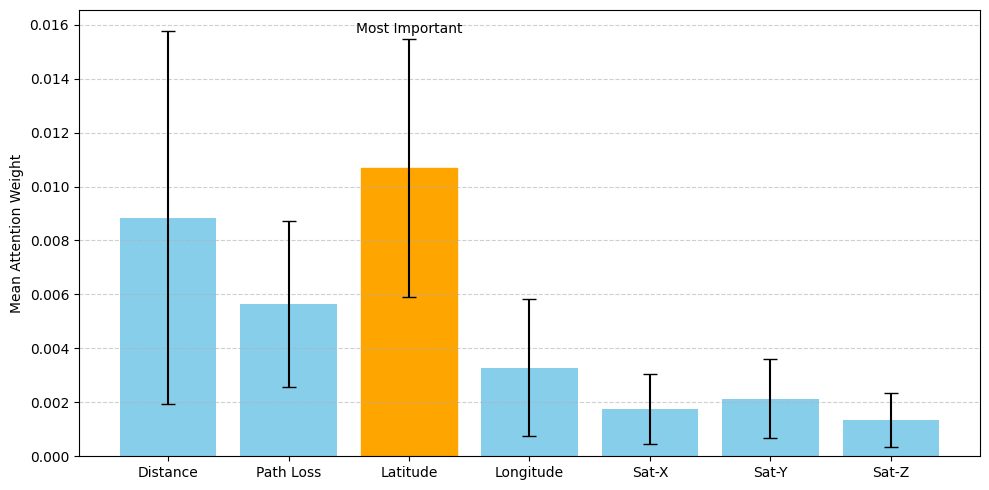}
    \caption{Mean attention weights across seven input feature categories with 95\% confidence intervals over 100 episodes.}
    \label{fig:attn_dist}
\end{figure}


Finally, we have looked into the behaviour of the internal attention mechanism to understand which features had the highest influence on the learned policy by the \acrshort{drl} agent. Figure~\ref{fig:attn_dist} shows the mean attention weights from seven main feature categories with 95\% confidence intervals. From the figure, we can observe that spatial features such as latitude and distance receive the largest weights, with path loss and longitude following closely, and the feature with lowest weight was the  $Z$-coordinate of the satellite. This is because, this feature does not change much since the orbit configuration has all satellites move at fixed altitude. Therefore, most of the useful geometry changes for link planning is in latitude, longitude, and slant distance rather than in $Z$ alone coordinate. The
results match what is observed in domain knowledge, in which latitude determines whether a terminal is in an equatorial or high-latitude region, and distance and path loss have more control on the instantaneous link budget. This finding shows that the proposed framework provides an interpretable view of the learned policy and shows that the \acrshort{llm}–\acrshort{drl} agent focuses on physical \acrshort{ntn} features more.


\section{Conclusion and Open Challenges}
\label{sec:conclusion}


In this paper, an  \acrshort{llm}–\acrshort{drl} framework for downlink resource allocation in \acrshort{leo} \acrshort{ntn}s was presented. The resource allocation problem was formulated as an  \acrshort{mdp} with \acrshort{llm} that translates network states into stratiges to be followed by the learning agent. The resulting framework improves sum rate, reduces outage probability, and maintains higher fairness compared to the traditional black-box \acrshort{drl} and classical heuristics. The framework has also been shown to provide interpretable policies that can be justified from service providers aspect and serves as initial step towards transparent AI control for \acrshort{ntn}s.


Several open challenges are identified. First, there are limits in theoretical guidance of \acrshort{llm} embedded in \acrshort{rl}. For example, how prompts affect convergence or training stability, and optimality guarantees. Second, scalability and implementation issues remain an open challenge. This paper methodology considered a moderate constellation with single beam per user, and one query per episode. Future works, could test large constellations, inter-satellite interference, as well as adding domain knowledge to the \acrshort{llm}. Finally, prompt engineering remain an open challenge, in which we have observed that \acrshort{llm}s can generate inconsistent responses. Domain knowledge, continual learning, and confidence aware mechanisms could help in minimizing these inconsistencies. Addressing these challenges will improve the progress towards realizing \acrshort{lam}–\acrshort{drl} in next-generation satellite resource management.


\section*{Acknowledgments}
We acknowledge the financial support provided by the Centre for Research Management, Sunway University, through the postdoctoral research scheme to the first author.

\section*{Conflict of interest}
The authors declare that there is no conflict of interest in this paper.
\bibliographystyle{elsarticle-num}

\end{document}